\documentclass{llncs}
\usepackage{llncsdoc}
\usepackage[utf8]{inputenc} % allow utf-8 input
\usepackage{booktabs}       % professional-quality tables
\usepackage{amsfonts}       % blackboard math symbols
\usepackage{nicefrac}       % compact symbols for 1/2, etc.
\usepackage{microtype}      % microtypography
\usepackage{graphicx}

\def\YH#1 {{\bf x}_{\it {#1}}}
\def\YI#1 {{\bf u}_{\it {#1}}}
\def\YO#1 {{\bf o}_{\it {#1}}}
\def\YT#1 {{{\bf u}^{\sf teach}_{\it {#1}}}}
\def\YHLIN#1 {{\bf x}_{{\sf lin}, \it {#1}}}

\def \ti {\it t}
\def \tanh {\sf tanh}

\def\WWHH {{\bf W}}
\def\WWIH {{\bf w}^{in}}

\newcommand{\appropto}{\mathrel{\vcenter{
  \offinterlineskip\halign{\hfil$##$\cr
    \propto\cr\noalign{\kern2pt}\sim\cr\noalign{\kern-2pt}}}}}

\author{Norbert Michael Mayer}
\institute{Department of Electrical Engineering and AIM-HI\\
           National Chung Cheng University,Taiwan,        
\email{mikemayer@ccu.edu.tw}}

\begin{document}

\title{Critical Echo State Networks that Anticipate Input using Morphable Transfer Functions}

\maketitle

\begin{abstract}
  The paper investigates a new type of truly critical echo state networks where individual transfer functions for every neuron can be modified to anticipate the expected next input. 
  Deviations from expected input are only forgotten slowly in power law fashion. The paper outlines the theory, numerically analyzes a one neuron model network
  and finally discusses
  technical and also biological implications of this type of approach.
\end{abstract}

\section{Introduction}
\label{sec:intro}
Recurrent neural networks (RNNs) with input are examples for non-autonomous dynamical systems. One fundamental property is their dependence on 
their initial states (i.e. the initial settings of the recurrent layer neurons) with regard to one given input sequence.
On one hand and for obvious reasons, networks that sensitively and for all future states depend on the setting of the initial state, will not work very well. On the other hand,
if the network forgets too fast information about the past, it essentially works in the same way as a feed-forward network, and if that is good enough 
for the given task it is much easier to replace the recurrent network with the feed-forward solution.
In the field of reservoir computing\cite{lukovsevivcius2009reservoir,schrauwen2007overview} and particular in the case of echo state networks (ESNs)\cite{jaeger1,jaeger2,jaeger2007special} much efforts have been undertaken in order to quantify to which extend an RNN is sensitive to the initial state.
As a result several methods exist to detect the fine line between network parameters that -- in combination with a given input sequence -- finally result in a forgetting of the past within the network
versus such parameter values for which essentially differences in the initial settings prevail in all future. More interestingly, heuristics show that parameter settings that are near the border line, 
however on the side of the forgetting type of networks, show the best performance for certain relevant tasks\cite{2005natschlaeger,criticalESNs2006,theobi2012}. These networks are called near critical networks. 
%, with an emphasis on {\em near}.
%\footnote{Also several algorithms to measure the distance to criticality have been proposed \cite{livi2016determination,wainrib2016local}.  } . One possible explanation for a better performance of the near critical network is that those networks account for a larger window of the recent input history than under-critical approaches.    
An important notice from experimental biology is that also the statistics of 
dynamics of neurons in brain slices hint towards a near critical or even critical tuning of biological neurons in the brain\cite{slices}.
Practical state of the art near critical networks usually require a certain margin towards the critical state because by design unexpected input deviations may push the state of the network over the critical point, in which case the  
performance deteriorates. In contrast to near critical networks, a relatively new study\cite{neco2015,myselfarxiv} brought up the idea to train the synaptic weights of the recurrent layer in the way that certain points (so-called epi-critical points, ECPs) within the transfer function are hit. If the network receives unexpected input these special points are missed and result in an under-critical behavior. Given an expected input the resulting network is tuned exactly to the critical point; other network features are power law forgetting of an unexpected input if it is succeeded by a sequence of expected input.  
Although that approach lines up a complete and new concept of designing critical ESNs, for practical purposes, there are still some problems. Most important, the proposed learning algorithm does not guarantee for a good performance of the network for many tasks. Different from that approach the present work does not apply learning to the input weights and the recurrent weights. Rather, it proposes adaptive transfer functions for each neuron where the ECPs are always shifted towards the next expected transition point.

\section{Echo State Networks and Criticality}
\label{sec:ESN}

The system is intended to resemble the dynamics of a biological 
recurrent neural network. We follow here the notation of J\"ager:
\begin{eqnarray}
\YH { lin, \ti} &=&   { \WWHH \YH { \ti-1 } } + \WWIH \YI { \ti } \label{xlin}\\
\YH { \ti} &=& \theta \left(   \YH { lin, \ti}     \right) = f(\YH{ \ti-1 } , \YI { \ti  } ),  
\label{esn_dyn} 
\end{eqnarray}
where $\YI { \ti } \in \mathbb{R}^{n}$ are items that form a left infinite time series that in total are called $\bar{\bf u}^{\infty}$
Supervised learning is done by linear regression using $\YH { \ti} $ as input\cite{jaeger1}, $\YH { t } \in \mathbb{R}^{k}$ represents activity in the hidden layer. 
$\WWHH$ and $\WWIH$ are matrices that represent (constant) synaptic weights. 
In principal, the complete time series $\YH { \ti } $ is determined by the tuple of the initial state $\YH { 0 } $ and a time series $\bar{\bf u}^{\infty}$. Comparing 
two time series 
%%\begin{eqnarray}
${\bf y}_{t} =f({\bf y}_{t-1} , \YI { \ti } )$ % \nonumber \\
and
${\bf x}_{t} =f({\bf x}_{t-1} , \YI { \ti } )$,
%\end{eqnarray}  
that start with any combination of two different ${\bf y}_{0} \neq {\bf x}_{0}$ one quantify how the difference develops over time.
Important for the definition of 
echo state networks is the concept of state contraction that is if
\begin{equation}
\lim_{t \rightarrow \infty} ||  {\bf y}_{t} - {\bf x}_{t} ||_2 = 0, \label{convergence}
\end{equation}
i.e. the Euclidean distance converges to zero.
In combination with the assumption that the processing of the neural network is acting in a time invariant manner, the concept has been named uniformly state contracting system\cite{jaeger1}.
Uniformly state contracting networks are echo state networks and thus capable to learn by linear regression.
There has been some confusion about the definition of uniformly state contracting networks. Some researches define it to describe the dynamics of a network with regard to a specific input sequence (cf. \cite{manjunath2013echo}). Within this paper networks are called uniformly state contracting only if for a given network the relation of eq. \ref{convergence} holds for {\bf any} input. Some calculus shows that a network with the dynamics of eq. \ref{esn_dyn} is always an ESN if the recurrent connectivity matrix is orthogonal ($\WWHH \in O(k)$) and the derivative of the transfer function is in the range $0 \leq \dot{\theta}(.) < 1$. Single inflection points, where $\dot{\theta}(.) = 1$ may be permissible
\cite{jaeger1,myselfarxiv}. These points are important in the following considerations, and are called epi-critical points (ECPs).
In analogy for calculating the Lyapunov exponent in autonomous dynamic systems one can define also a Lyapunov exponent for ESNs with regard to a certain input sequence  $\bar{\bf u}^{\infty}$ 
\cite{wainrib2016local,myselfarxiv}
\begin{equation}
\Lambda_{\bar{\bf u}^{\infty}} = \lim_{|{\bf y}_0-{\bf x}_0|\rightarrow 0 } \; \; \lim_{t \rightarrow \infty} \frac{1}{t} \log \frac{||{\bf y}_t-{\bf x}_t||_2}{||{\bf y}_0-{\bf x}_0||_2}.
\label{lyapunov_eq}
\end{equation}
If the Lyapunov exponent for all  $\bar{\bf u}^{\infty}$  is negative, the network is shown to be uniformly state contracting and thus is an ESN. If the Lyapunov exponent for any input time series is positive the network is not an ESN.
In addition, it is worthwhile to introduce the following definition: A network that for some input sequences has a Lyapunov exponent of zero but for which the eq. \ref{convergence} still holds for any input sequence shall be called
a {\bf critical} ESN. According to this definition an ESN is critical with respect to a particular input sequence.
%\footnote{For finite size, ESNs that are critical to all possible input sequences seem hardly possible. One reason is the limited information content within the reservoir, which contradicts with the idea of power law forgetting w.r.t. all possible input sequences.  }. 
Technically, this can be achieved by training the network towards a setting where for some input sequences
%\begin{equation}
$\dot{\theta}(\YH { lin, \ti } ) = 1$.
%\footnote{This equation is true for orthogonal recurrent connectivity. 
%The larger set of connectivity using normal matrices the corresponding formula is 
%\[
%\sigma_{\max} \left(\WWHH \right) \cdot \dot{\theta}(\YH { lin, \ti } ) = 1, \nonumber
%\]
%where $\sigma_{\max}$ is here the largest singular value of $\WWHH$. 
A deeper insight into the theory behind that formula can be found in \cite{neco2015,myselfarxiv}.%}
%\end{equation}
That is to direct the input to those single inflection points of the transfer function. 
%%Conceptually, the key idea is to achieve the equation above in the case that an expected, i.e. a very boring, or usual input is presented to the network.

\begin{figure}[t]
\begin{center}
\includegraphics[width=0.25\paperwidth]{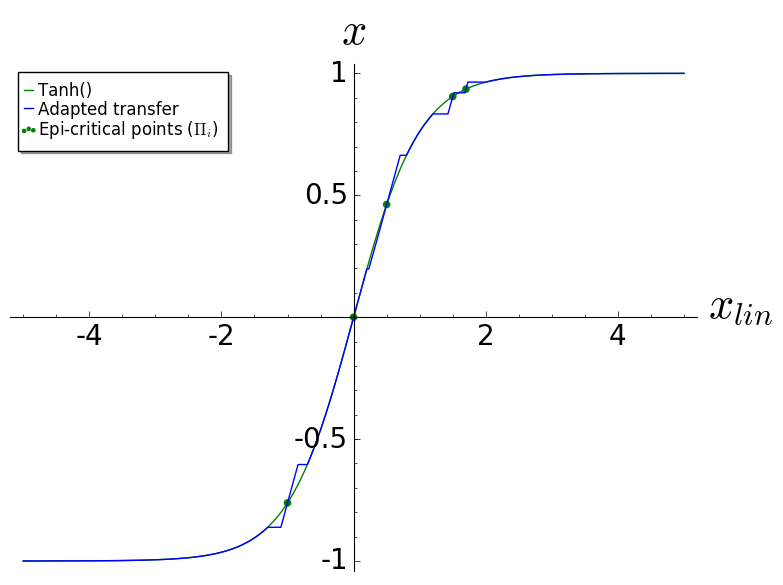} %\hskip 0.35cm
\includegraphics[width=0.25\paperwidth]{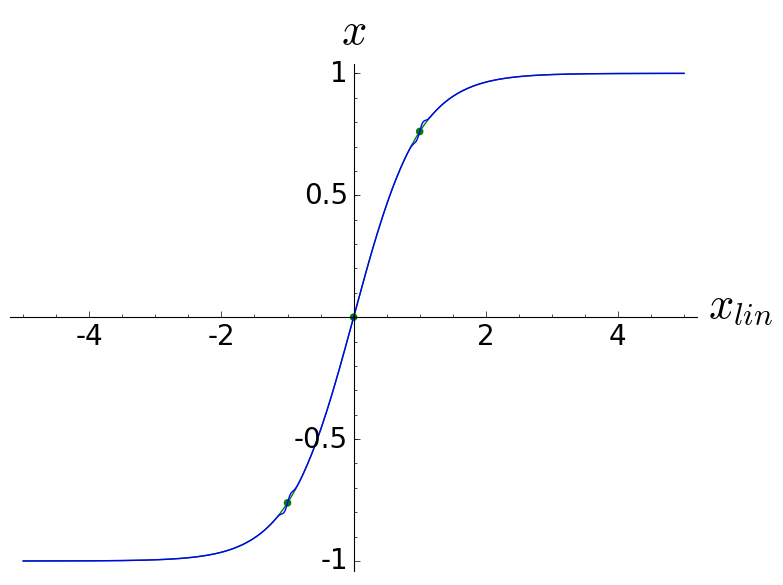}
\end{center}
\caption{\label{tailored_transfer} 
The plots show examples of two versions of a transfer function $x=\theta(x_{lin})$ according to eqs. \ref{adapt_trans1}  -- \ref{adapt_trans2} as blue curve where the ECPs can be organized in an adaptive way. The green curve depicts
the underlying $\tanh$ function on which all ECPs are located.   
Green dots indicate ECPs for this particular example. The version at the right side is the version that is used in sect. \ref{sec:Transfer}. Here areas of the transfer function where $\dot{\theta}=0$ have been avoided
which leads to better results in the following graphs. Since the derivative of $\tanh(0)$ is $1$ the point $(0,0)$ is marked as an additional ECP in both plots.  
}
\end{figure}

\section{Adaptive transfer functions}
\label{sec:Transfer}

Instead of implementing plasticity on $\WWHH$ and $\WWIH$, the proposal for the current work is to implement the plasticity on the shape of the transfer function $\theta$. Assuming that each neuron has an intrinsic mechanism to predict several possible 
values of $x_{ lin, \ti} $ (that is one item in the vector $ \YH { lin, \ti} $) the prediction happens before the input $\YI { \ti } $ is perceptible to the neuron (compare \cite{BIOADIT2004,NIPS+2006} for another scenario for self-prediction in ESNs). The neuron does not need to restrict itself to one prediction, instead a list of those values are possible. So, the ECPs $\Pi_i$ should be shifted towards the predicted values of the linear response. The transfer function around an ECP ($\Pi_i \approx x_{lin}$)  can thus be defined as
\begin{equation}
\theta(x_{lin}) = {\tanh}(x_{lin}-\Pi_i)+{\tanh}(\Pi_i), \label{adapt_trans1}
\end{equation}
or, else
\begin{equation}
\theta(x_{lin}) = {\tanh}(x_{lin}).  \label{adapt_trans2}
\end{equation}
The detailed arithmetic of such a function is more complicated and would take much space to be detailed out here (for example it needs to be defined what is the function value between two values that are located nearby each other.) 
Fig. \ref{tailored_transfer} depicts two possible resulting transfer functions. The transfer function is designed in the way that
%%\begin{equation}
$\dot{\theta} (\Pi_i ) = 1$.
%%\end{equation}
%%Some may see the idea of adaptive transfer functions as a new version of intrinsic plasticity \cite{triesch2005gradient,steil2007online}.

\section{Synthetic one neuron reservoirs}

\begin{figure}[t]
\begin{center}
\includegraphics[  width=0.25\paperwidth]{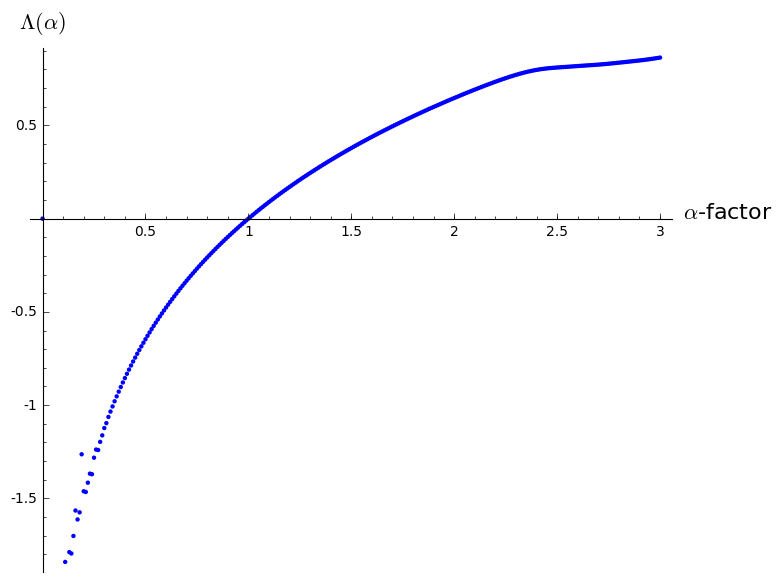}
\includegraphics[  width=0.25\paperwidth]{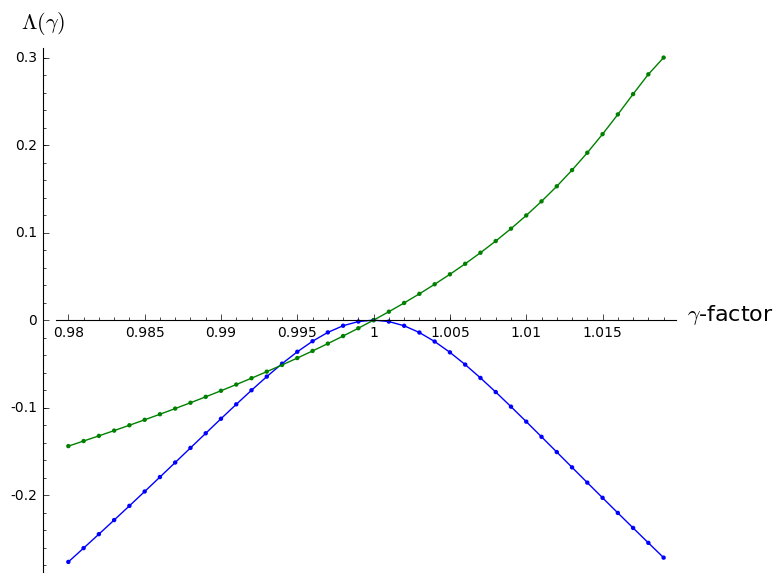}
\end{center}
\caption{\label{fig:lyapunov} Left: Depicted is the Lyapunov exponent for the example system of eq. \ref{alternating} for different values of $\alpha$. At $\alpha=1$ the Lyapunov exponent crosses zero if the input sequence is the expected alternating sequence of 1s and -1s.
Right: Lyapunov exponents for two one neuron networks: In both cases the amplitude of the alternating input $u_t$ is varied in a series of measurements of the Lyapunov exponent.
In the case of the network according to eq. \ref{alternating} (blue) one can see that the Lyapunov exponent never becomes larger than 0. In the case of eq. \ref{alternating_old} (green) positive Lyapunov exponents occur
if the amplitude of the input is larger than the critical value. 
}
\end{figure}

\begin{figure}[t]
\begin{center}
\includegraphics[  width=0.185\paperwidth]{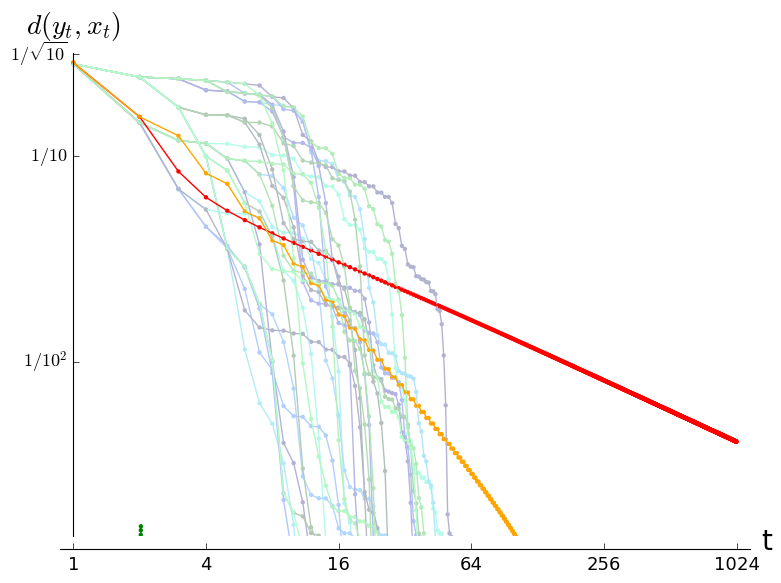}
\includegraphics[  width=0.185\paperwidth]{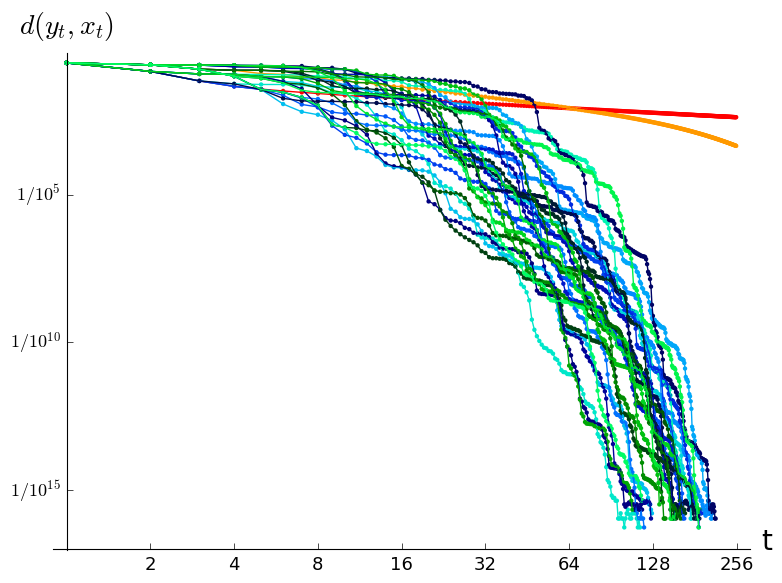}
\includegraphics[  width=0.185\paperwidth]{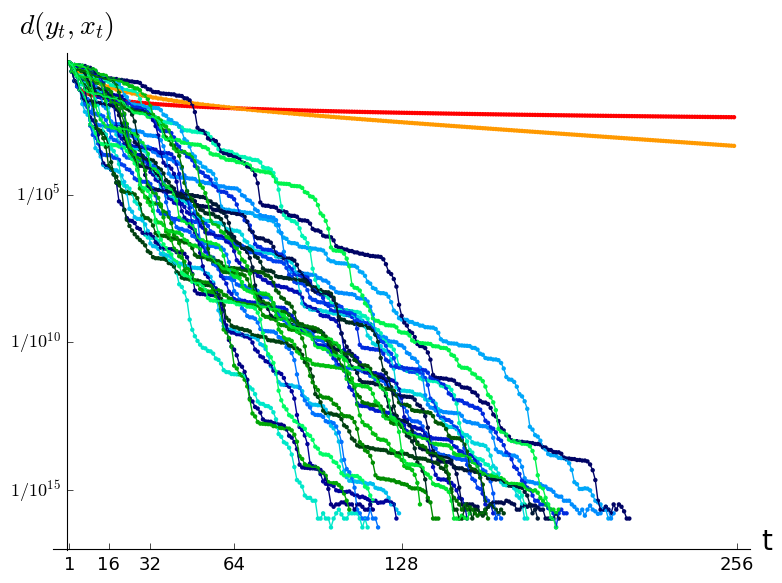}
\end{center}
\caption{\label{fig:powerlaw_result} The graphs depict different versions of the same data. Each red curve is the forgetting curve of the initial difference between 2 networks if the input sequence is alternating between $1$s and $-1$s. The orange curves depicts the forgetting curve for constant input with amplitude $1$. The other curves show several iid. random sequences of $1$ s and $-1$s with equal probability. The left plot is a log-log plot with a focus on the alternating and constant input. The red curve converges towards a straight line, which indicates an underlying power law of this data. The curve resulting from constant input shows large values even at later time steps. However, the convergence appears to be faster than a power law, hence the curve bends toward the bottom. Finally random input shows the fastest convergence. From the middle and right side graphs clearly indicate that all except for the alternating input show (roughly) an exponential decay. Both the middle and the right side plot show a scale down to $10^{-15}$, which is about the limit of precision of double precision floating point numbers. Once a difference between two initial states reaches zero it is beyond the logarithmic scale and not plotted anymore. So, the curve ends at that iteration.  }
\end{figure}

In order to illustrate the proposed methodology we designed an a reservoir with one neuron that expects an alternating input of $u_t=1$s and $-1$s. The following update equation can then be used
\begin{equation}
x_{t} = \theta \left( - \alpha x_{t-1} + (1- \alpha \, {\tanh}(1)) u_t \right), \label{alternating}
\end{equation}  
where the factor $\alpha$ takes the role of a $\WWHH \in R^{1\times1}$ matrix and the factor $(1- \alpha \, \tanh(1))$ the role of $\WWIH$. Note that if $\alpha=1$ one may call $\WWHH=O(1)$ a one dimensional orthogonal
matrix. Since for any $\alpha$ and in case the network receives the expected alternating input the linear response converges to $x_{lin}= \pm 1$. So, two ECPs may be used
$\Pi_0=-1$ and $\Pi_1$=1. With regard to the resulting transfer function (that includes the ECPs)  one can now measure the Lyapunov exponent according to eq. \ref{lyapunov_eq} and for differing values of $\alpha$. Fig. 
\ref{fig:lyapunov} (left) depicts the results. One can see that although independent from $\alpha$ for the predicted input the dynamics are always alternating $1$s and $-1$s, this dynamic is only stable for the range of $\alpha$ between $0$ and $1$. In this range the network is an echo state network, that is under-critical if $\alpha<1$. Further numerical tests and also theoretical considerations show that at the point $\alpha=1$ the network is still an ESN, however critical. For values of $\alpha>1$ the network is not an ESN anymore. The purpose of this work is to propose ESNs of $\alpha=1$ as optimal critical ESNs.
For comparison one may consider a one neuron version of a traditional near edge of chaos approach which basically relate to 
the common experience that the given theoretical limits for the ESNs can be significantly overtuned for many practical time series. 
Those overtuned ESNs in many cases show a much better performance than those that actually obey Jaegers initial limit. 
So recently researchers came up with theoretical insights with regard to ESNs that are subject to a network and a particular input statistics \cite{manjunath2013echo} which fundamentally relate 
a network {\bf and} an input statistics to a limit. 
One might assume that those approaches show similar properties as the one that has been presented above. 
However, for a good reason those approaches all are coined as 'near edge of chaos' approaches.
In order to illustrate the problems that arise from those approaches one may consider what happens if those overtuned ESNs are set exactly to the critical point. Here, just for the general understanding one may
consider again a one neuron network and a $\tanh$ as a transfer function, so
\begin{equation}
x_{t} = {\tanh}(-b x_{t-1} + u_t). \label{alternating_old}
\end{equation}
Note that the ESC limit in outlined above requires that the recurrent connectivity should be $b<1$. 
One can now take the input time series from the previous section
%%\begin{equation}
$u_t=(-1)^t \pi/4 \label{input_ts}$.
%%\end{equation}
Slightly tedious but basically simple calculus results in a critical value of $b\approx2.344$ for the input time series, where 
$x_t \approx (-1)^t \times 0.757$. For the following results the value of $b$ is always set to the critical value. 
In this situation one can test for convergence of two slightly different initial conditions and one can get a power law decay of the difference. 
However, setting up the amplitude of the input just a tiny bit higher is going to result in two diverging time series $x_t$ and $y_t$. So: If the conditions of the ESN are chosen to be 
exactly at the critical point it is possible that a not trained input sequence very near to the trained input sequence can turn the ESN in a state where J\"ager's echo state condition is not fulfilled anymore,
i.e. that Lyapunov exponent is positive for the given network in combination with these input sequences. 
%\begin{figure}[t]
%\begin{center}
%\includegraphics[  width=0.3\paperwidth]{critical_both.png}
%%\includegraphics[  width=0.32\paperwidth]{critical_old_ver.png}
%\end{center}
%\caption{\label{fig:lyapunov2} }
%\end{figure}
In order to illustrate this difference numerical experiments (cf. Fig. \ref{fig:lyapunov} right side) have been done were both the network according to 
equation \ref{alternating} and of eq. \ref{alternating_old} receive input with a slightly higher or lower input amplitude, i.e. an input sequence $\tilde{u}_t = \gamma \cdot u_t$ is perceived, where $\gamma$ is a constant factor and 
and $u_t$ in both cases is the expected input that produces the critical behavior. Here, the amplitude $\gamma$ is used as an example as an arbitrary continuous parameter that defines properties of the input sequences. 
If $\gamma$ is equal to one, the resulting input sequence for both, the example of eq. \ref{alternating} and eq. \ref{alternating_old}, results in a critical dynamics with a Lyapunov exponent of $1$. The difference between the two networks is that in the case of eq. \ref{alternating_old} positive Lyapunov are possible for $\gamma$-factors larger than one whereas in the case of the proposed network for any input sequence the Lyapunov coefficient is smaller or equal to one. 
This means that the network of eq. \ref{alternating_old} is not an ESN according to the definition given in sect. \ref{sec:ESN}, while the proposed network is an ESN if the convergence condition of eq. \ref{convergence} holds. 
Analytic calculus \cite{myselfarxiv} shows that in the critical case the nature of the transfer function determines if a network is an ESN or not. 
Fig. \ref{fig:powerlaw_result} depicts the convergence process of two exemplary start values at the critical state and the one neuron network of eq. \ref{alternating} and compares the results of the expected alternating 
input ($1$s and $-1$s) with constant input of the same amplitude and an iid. random set of $1$s and $-1$s. In the first case double log plots reveal that the vanishing follows a power law, i.e. forgetting is slow. Thus, we get
\begin{equation}
d(x_t, y_t) \appropto t^{-c_a}, 
\end{equation}
with a constant $c_a$.
The other types of input statistics result in faster forgetting. Here, every input value may be seen as an event that demands memory capacity. The result is effectively an exponential forgetting
i.e.
\begin{equation}
d(x_t, y_t) \appropto {c_b}^t, 
\end{equation}
with a constant $c_b$. Which is the same result the one would also expect for all memory decay in under-critical networks. Exponential decay appears as a straight line in {\bf semi-logarithmic} plots.
The single neuron network simulations have been done by using double precision floating point variables, i.e. in 64 bits. 
%according to IEEE 754-2008, which formalizes a structure of three parts, 
%sign (1 bit), exponent (11 bits)  and mantissa (51 bits). Obviously, the theoretical limit information capacity of a one neuron reservoir is 64 bits. 
Since the experimental setting in fig. \ref{fig:powerlaw_result} organizes the 
initial difference between the 2 networks in an identical way as the randomness of the following inputs (that are identical to both networks) one would expect that the differences vanish over the time of 64 iterations. So one 
expects that the difference between the 2 network vanishes roughly in about 64 iterations. 
Considering the results of  fig. \ref{fig:powerlaw_result} one can see that indeed the difference vanishes in about 64 to 200 iterations.The fact, that the 
forgetting process is slower than 64 iterations may indicate that several variant input histories can result in the same identical reservoir state.

\section{Discussion}

%In conclusion there are several aspects that can be brought up here. The first group of the following conclusions is related to technical issues, the second deals with biological implications. 

%{\bf Technically}, 

ESNs can be tuned to the critical value on the spot. At the same it can be guaranteed that no input can
push the network over the permissible limit. The setting of the ECPs leads to new insights into the network dynamics and relate those to information theory. If the next input is predictable, the next state of the network 
is going to hit one ECP exactly. One may interpret the resulting network in the way that predictable input  is always directed to the ECPs and in this way prevented from consuming too much space (i.e. entropy) in the reservoir. Instead, {\bf deviations} from predicted input materialize in the reservoir as distances to the ECPs. These deviations prevail than in a power-law fashion.
This is true for both the present approach and the approach proposed in \cite{neco2015}. 
Different from \cite{neco2015} the approach here focuses on a adaptive transfer function. Overall there is very limited literature about adaptive transfer functions in neural networks (e.g. \cite{wang2006general} ). With regard to reservoir computing investigations into adaptive transfer functions may be promising. 
In the present approach one target of the investigation was to find a way of training where the position of the transition point 
$\theta({\bf x}_{lin,t}) $ was unchanged, only the environment around it was transformed in the way that $\dot{\theta}({\bf x}_{lin,t}) = 1 $. This method in some sense changes the topology of the reservoir: 
By design, in every transition reservoirs loose information about previous inputs, however this information loss is not homogeneous and independent from the input time series. Rather it varies depending on features of the network, on the current input value and other parameters.
Using the method of ECPs the reservoir transforms then into a magnifying glass around those predicted states, which allows the network to look deep into the past if the incidence of aberrations from the predicted values are rare. So, aberrations from the predicted states can leave traces in the reservoir for very long times -- if they are rare. In this sense the input-driven network turns into an {\bf event-driven} network, i.e. a system that reacts strongly on an unpredicted event in contrast to the everyday and usual input. Or, to put it in other words a lossy memory compression of a sliding window with an infinite but more and more lossy reproducibility of the far past.
{\bf Acknowledgements.} This manuscript has been posted at arxiv.org. % \cite{myselfarxiv}
The authors thanks MOST of Taiwan for financial support and O. Obst for all his help.

\tiny

\bibliographystyle{unsrt} 
\bibliography{proof}

\end{document}